\documentclass[letterpaper, 10 pt, conference]{ieeeconf}  %

\usepackage[T1]{fontenc}
\usepackage[utf8]{inputenc}
\usepackage[inkscapearea=page]{svg}
\usepackage{balance}    %
\usepackage{soul}  %
\usepackage{textpos}    %
\usepackage{rotating}   %
\usepackage[normalem]{ulem}
\usepackage{xcolor}

\IEEEoverridecommandlockouts                              %
\usepackage[l2tabu,orthodox]{nag}

\usepackage[
    backend=bibtex8,
    style=ieee,
    sorting=none,
    natbib=true,
    doi=false,
    isbn=false,
    url=false,
    eprint=false,
    maxcitenames=1,
    mincitenames=1
]{biblatex}

\usepackage[pdftex,colorlinks]{hyperref}
\usepackage[printonlyused]{acronym}

\usepackage{siunitx}
\usepackage[all]{nowidow}

\usepackage{lipsum}

\usepackage{xspace} %
\newcommand{\ie}{i.e.,\xspace{}}
\newcommand{\eg}{e.g.,\xspace{}}

\usepackage{epstopdf}

\usepackage{import}

\graphicspath{{./latexGoodPractices/}}

\usepackage{subcaption}

\usepackage{booktabs}

\usepackage{tabularx}
\usepackage{multirow, multicol}

\usepackage{amssymb,amsfonts,amsmath,amscd}

\usepackage{bm}

\newcommand{\bbm}{\begin{bmatrix}}
\newcommand{\ebm}{\end{bmatrix}}

\usepackage{etoolbox}
\makeatletter
\patchcmd{\@makecaption}
  {\scshape}
  {}
  {}
  {}
\makeatother

\acrodef{RoIs}{regions of interest}
\acrodef{BEV}{bird's-eye view}
\acrodef{RPN}{Region Proposal Network}
\acrodef{NMS}{non-maximal suppression}
\acrodef{SSD}{Single-Shot MultiBox Detector}
\acrodef{IoU}{intersection over union}
\acrodef{mIoU}{mean intersection over union}

\title{\LARGE \textbf{
    MaskBEV: Joint Object Detection and Footprint Completion for Bird's-eye View 3D Point Clouds
}}

\author{William Guimont-Martin$^{1}$, Jean-Michel Fortin$^{1}$, François Pomerleau$^{1}$, Philippe Giguère$^{1}$%
    \thanks{\raggedright$^{1}$ The authors are with Northern Robotics Laboratory, Université Laval, Québec City, Canada,
    {\texttt{\small{william.guimont-martin.1@ulaval.ca, philippe.giguere@ift.ulaval.ca}}}}%
    \thanks{$*$ The source code for this paper is available here: \href{https://github.com/norlab-ulaval/mask_bev}{https://github.com/norlab-ulaval/mask\_bev}}
    \thanks{$**$ © 2023 IEEE. Personal use of this material is permitted. Permission from IEEE must be obtained for all other uses, in any current or future media, including reprinting/republishing this material for advertising or promotional purposes, creating new collective works, for resale or redistribution to servers or lists, or reuse of any copyrighted component of this work in other works.}
}

\usepackage[switch]{lineno}

\begin{document}

\vspace{-7mm}
\maketitle
\thispagestyle{empty}
\pagestyle{empty}

\begin{abstract}

Recent works in object detection in LiDAR point clouds mostly focus on predicting bounding boxes around objects.
This prediction is commonly achieved using anchor-based or anchor-free detectors that predict bounding boxes, requiring significant explicit prior knowledge about the objects to work properly.
To remedy these limitations, we propose MaskBEV, a \ac{BEV} mask-based object detector neural architecture.
MaskBEV predicts a set of \ac{BEV} instance masks that represent the footprints of detected objects.
Moreover, our approach allows object detection and footprint completion in a single pass.
MaskBEV also reformulates the detection problem purely in terms of classification, doing away with regression usually done to predict bounding boxes.
We evaluate the performance of MaskBEV on both SemanticKITTI and KITTI datasets while analyzing the architecture advantages and limitations.

\end{abstract}

\section{Introduction}

Object detection in 3D point clouds is crucial to many applications of robotics and autonomous vehicles.
Point clouds, captured by sensors such as LiDARs, provide accurate 3D information about the system’s surroundings.
However, it is more difficult to process them with deep neural networks in order to extract actionable semantic information. 
Indeed, point clouds, unlike images that are a dense regular grid of pixels, are irregular, unstructured and unordered \citep{Bello2020}. 
Moreover, LiDAR point clouds suffer from multiple types of occlusion and signal miss \cite{Xu2022}.
One type of occlusion is \emph{external-occlusion}, which is caused by obstacles blocking the laser from reaching the objects.
\emph{Self-occlusion} happens when an object's near side hides its far side.
It is inevitable and will affect every object in a LiDAR scan.
\emph{Signal miss} can be caused by reflective materials reflecting the laser beam away from the sensor or by low reflectance.
This often leads to objects appearing incomplete in LiDAR scans.

As such, 3D object detection models need to take into account these particularities of LiDAR point clouds to achieve good accuracy.
This is especially true for object detection in an autonomous vehicle context, where only the part of the object facing the LiDAR will be visible.
This makes objects incomplete and harder to detect \citep{Xu2022}.

Most recent 3D object detection architectures are based on predicting a set of 3D bounding boxes around objects. 
These methods can be separated into two main categories: \emph{anchor-based} and \emph{anchor-free} detectors. 
Anchor-based approaches use a predefined set of box proposals for potential object localization, along with their dimensions \citep{Chen2019, Shi2021, Deng2020, Mao2021, Xu2022, Lang2018, Mao2021, Zheng2021}. 
A notable drawback of this type of approach is the need to employ a large number of anchors, which are regulated by numerous hyperparameters related to their locations and sizes \citep{Law2018}. 
Anchor-free detectors instead directly predict the object's location and shape using keypoints from feature maps \citep{Zhou2019, Law2018, Wang2022, Hu2022, Chen2020, Yin2021, Li2021, Sun2022, Yang2020, Li2021, Zhou2019, Ma2022, Zou2022}, such as the corners of the bounding box, or the object's center and size.
Anchor-free detectors often heavily depend on post-processing steps such as \ac{NMS}, thresholding, and max pooling.
These hand-crafted components (e.g., anchors in an anchor-based detector and \ac{NMS}) explicitly encode prior knowledge about the objects \citep{Carion2020} and thus require careful tuning.

\begin{figure}[tbp]
    \centering
    \includegraphics[width=\columnwidth]{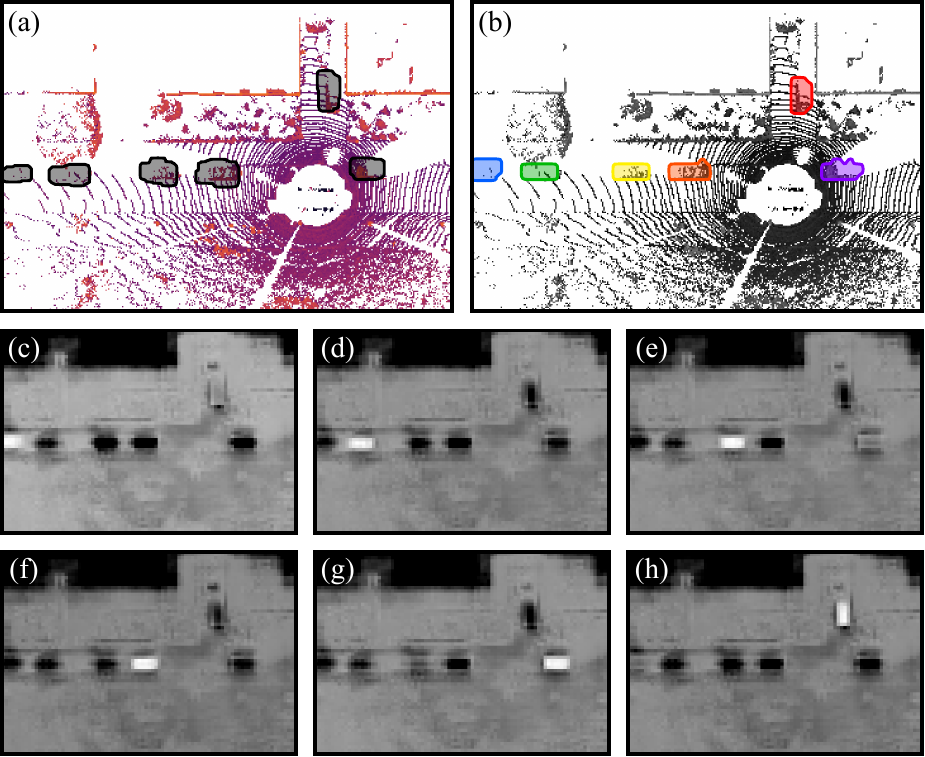}
    \caption{\textbf{Mask prediction from MaskBEV.}
    (a) \ac{BEV} of a point cloud from SemanticKITTI's validation set.
    We overlay the ground truth masks capturing each object's footprint on the point cloud.
    (b) Mask predictions from MaskBEV overlayed on the input point cloud, each instance is predicted on a different mask, here shown in different colors.
    (c-h) Mask predictions before applying the sigmoid and thresholding operation. 
    We clipped and normalized the raw predictions to make them easier to visualize. 
    We can notice the black outlines of cars not predicted by a particular mask.
    This means that each query token specializes in detecting one instance while suppressing the others.}
    \label{fig:mask-log}
    \vspace{-1.5em}
\end{figure}

To address these limitations and to further adapt object detection architectures to the particularities of point clouds, we propose MaskBEV.
It is a \acf{BEV} mask-based 3D object detection neural architecture.
Instead of predicting bounding boxes, it predicts a \ac{BEV} instance mask capturing the footprint of each object as shown in \autoref{fig:mask-log}, along with their class.
Our approach allows us to \emph{simultaneously detect and perform object completion}, making our detection framework robust to occlusions and signal misses.
Importantly, our goal is to introduce a new 3D object detection paradigm based on masks instead of bounding boxes, as opposed to simply demonstrating superior accuracy over state-of-the-art models. 
In short, our contributions are:
\begin{itemize}
    \item a novel architecture, MaskBEV, demonstrating the advantages of 3D object detection based on masks;
    \item an algorithm to generate mask labels from both bounding boxes and semantic labels; and
    \item an evaluation of our approach for vehicle detection on KITTI  \citep{Geiger2013} and SemanticKITTI \citep{Behley2019} datasets.
\end{itemize}

\section{Related Work}

The prevailing approaches for detecting 3D objects involve predicting a set of bounding boxes.
These methods fall into two primary types: anchor-based and anchor-free detectors.
After highlighting the challenges of both these methods, we review mask-based detection.

\subsection{Anchor-based Detection}

Anchor-based detection methods rely on a predefined set of box proposals -- anchors -- to predict the location and dimensions of objects. 
The use of anchors comes in two flavors: \emph{single-stage}  and \emph{two-stage} methods.

Single-stage methods regress 3D bounding boxes in a single pass using anchors.
These anchors serve as reference points, as the final bounding box geometry will be regressed as offsets from these \citep{Liu2016}.
In the context of 3D object detection, the anchors usually encode prior knowledge about the possible object locations per class, their dimensions, their offset with respect to the ground, and their viewing angle \citep{Lang2018}. 
Multiple solutions have been proposed following this approach, namely VoTr-SSD \citep{Mao2021}, PointPillars \citep{Lang2018}, 3DSSD \citep{Yang2020},  SE-SSD \citep{Zheng2021}, and GLENet \citep{Zhang2022}.

Two-stage methods are largely based on Faster R-CNN \citep{Ren2015} which starts with a first stage where \ac{RoIs} are extracted from the input scene. 
This is accomplished by a \ac{RPN}, which uses anchors to define putative regions where an object instance might be present.
Based on this principle, methods such as Fast Point R-CNN \citep{Chen2019}, PV-RCNN++ \citep{Shi2021}, Voxel R-CNN \citep{Deng2020}, and BtcDet \citep{Xu2022}, work by using sub-networks to propose \ac{RoIs} directly in point clouds.
This allows the extraction of region-wise information that is then classified and used to regress bounding boxes. 
This usage of both classification and regression is typical for bounding box predicting methods.

While achieving competitive performance in 3D object detection, anchor-based methods inherently suffer from several drawbacks.
First, these methods require a large number of anchor boxes to cover possible object locations and sizes.
As such, it introduces numerous hyperparameters to be tuned, namely the number of boxes, their sizes, and aspect ratios \citep{Law2018}.
Second, the detections are predicted via regression with respect to these anchors.
Therefore, the quality of anchors -- how well they are aligned with objects -- has a significant impact on model performances \citep{Carion2020}.
Third, it complexifies the detection pipeline by adding a proxy task, the bounding box detection, compared to a single, unified model that directly predicts the presence or absence of an object in a particular location.
Fourth, bounding boxes might also contain parts of other objects.
This is particularly problematic if the anchors are axis-aligned, as it will have difficulties with diagonal elongated objects \citep{Fortin2022}, such as large vehicles.

Our MaskBEV approach, in contrast, directly predicts masks instead of bounding boxes, circumventing the need for anchors.
Consequently, it does not require any prior knowledge about the location and size of objects, while reducing the number of hyperparameters to tune.
Also, we reformulate detection purely in terms of classification without resorting to regression, which is often considered problematic \citep{Zhang2016}.

\subsection{Anchor-free Detection}

Anchor-free detection predicts an object's location and size using keypoints locations around objects.
CenterNet \citep{Zhou2019}, which detects objects in images by their center, inspired many subsequent 3D object detection architectures such as CenterNet3D \citep{Wang2022}, AFDetV2 \citep{Hu2022} HotSpotNet \citep{Chen2020}, CenterPoint \citep{Yin2021}, SWFormer \citep{Sun2022} and MGAF-3DSSD \citep{Li2021}.
Other methods instead predict the corners of 3D bounding boxes \citep{Ma2022} or points defining the characteristic L-shape of vehicles viewed from the side in a bird's-eye view point cloud \citep{Zou2022}.

Anchor-free detection avoids the limitation of anchor-based detection.
Instead, they often depend on complex post-processing steps to extract peaks in feature heatmaps (\eg{} \ac{NMS}, thresholding and max pooling) and rules to assign ground truth to predictions to ensure the quality of the predictions \citep{Carion2020}.
These processes improve the models' prediction and include additional design decisions that need to be tuned.
MaskBEV can side-step the need for such post-processing and complex ground truth assignation rules with the help of mask-based detection. 

\subsection{Mask-based Semantic and Instance Segmentation}

MaskFormer \citep{Cheng2021} and Mask2Former \citep{Cheng2022} are two recent semantic and panoptic segmentation models, designed for computer vision.
MaskFormer's insight, from which Mask2Former builds onto, is to consider this segmentation task as a mask classification instead of a per-pixel classification.
Pixel classification seeks to predict the class of each pixel of an image.
In contrast, mask classification predicts a set of binary masks, \ie{} bundles of pixels or points, and a class is predicted for each mask.
Ground truths are uniquely assigned to predictions using bipartite matching in a permutation-invariant manner.
This differs from methods often used in anchor-free methods that match predictions to the closest ground truth or to predictions with an \ac{IoU} above a user-defined threshold.
Masks with no matches are assigned a special \texttt{no\_object} class label.
Bipartite matching encourages predictions to be unique, thus eliminating the need for post-processing such as \ac{NMS} \citep{Carion2020}.

MaskRange \citep{Gu2022} reuses the same meta-architecture as Mask2Former but applies it to range images for semantic segmentation.
Mask3D \citep{Schult2022}, SPFormer \citep{Sun2022SP}, and MaskPLS \citep{Marcuzzi2023} all adapt the MaskFormer meta-architecture to produce point masks, \ie{} a point-wise binary mask for each instance.
Other approaches, not based on Mask2Former, also leverage mask prediction to do instance segmentation; 3D-BEVIS \citep{elich20193d}, 3D-BoNet \citep{Yang2019} and DyCo3D \citep{He2020} show the potential of mask-based detection for instance segmentation in LiDAR data.
However, they developed their approach for reconstructed indoor point clouds, and thus require point clouds with less occlusion than single-pose LiDAR scans.

Our MaskBEV leverages mask predictions to detect vehicles in LiDAR point clouds.
Consequently, instead of predicting a set of bounding boxes around each vehicle, MaskBEV predicts a bird's-eye view mask for each detected instance.
This allows MaskBEV to do away with the post-processing steps required by anchor-based methods, and to be more robust to occlusion inherent to single-pose LiDAR scans.
Moreover, the use of \ac{BEV} instance masks allows us to do object completion simultaneously with object detection, thus finding the complete footprint of a vehicle.
Masks also show themselves to be more flexible than bounding boxes: they allow capturing objects of any geometry, not just rectangular objects.
MaskBEV also eliminates the need for prior knowledge about objects' location and dimensions to be baked into the network (\eg{} anchors, post-processing steps); everything is learned directly from data.
Mask-based object detection also removes the need for regression and instead reformulates the instance detection task purely in terms of classification.

\section{MaskBEV}

\begin{figure}[htbp]
    \vspace{5pt}
    \centering
    \includegraphics[width=\columnwidth]{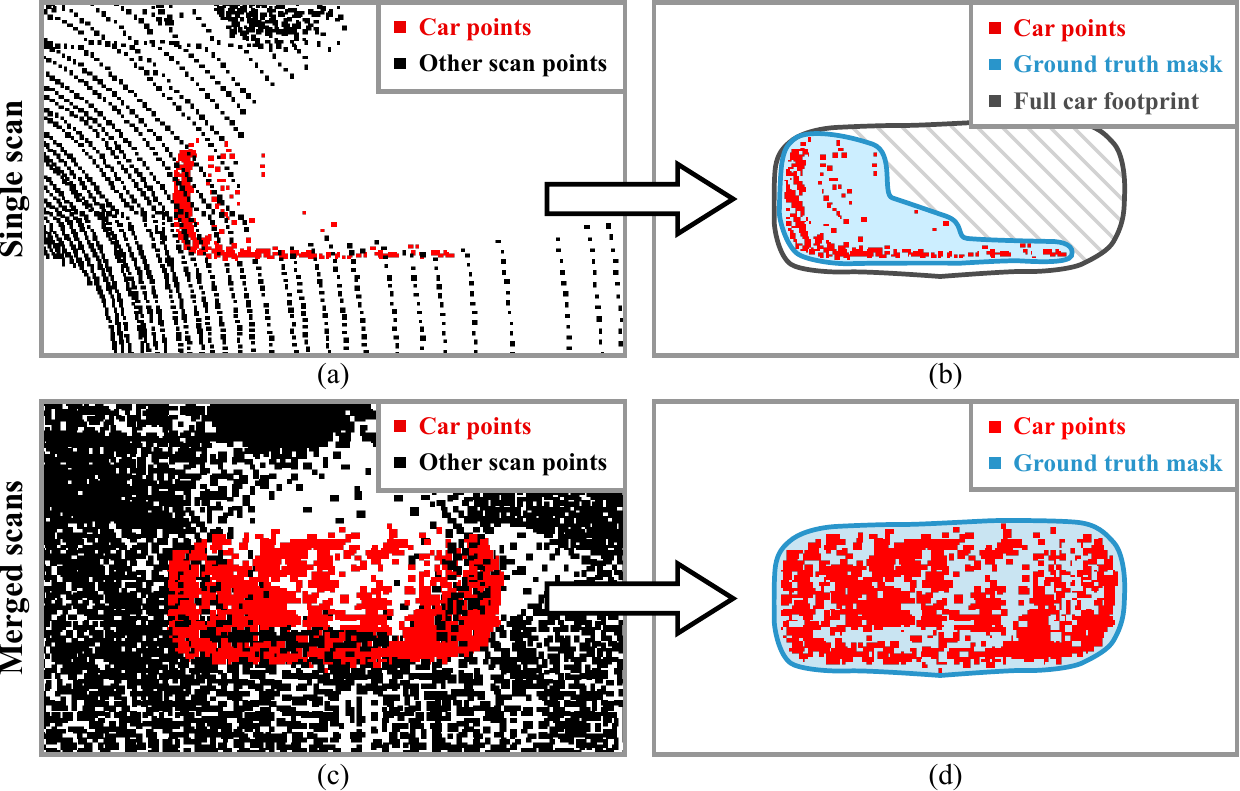}
    \caption{\textbf{Mask generation from instance label.} 
    (a) Single scan point clouds only show the surface of the object that directly faces the LiDAR.
    (b) The mask generated from a single scan is partial and does not represent the complete footprint of the vehicle.
    (c) Using merged sequential LiDAR scans, it is possible to gather points from all around a vehicle.
    (d) Masks produced from the constructed map are complete, \ie{} represent the entire footprint of the vehicle.}
    \label{fig:mask-completion}
    \vspace{-1.5em}
\end{figure}

MaskBEV is a novel object detection and footprint completion architecture for LiDAR point clouds using \ac{BEV} masks.
It is based on transformer networks in order to process global information by cross- and self-attention over the entire \ac{BEV}.
At the moment, we strictly target vehicles, as transformer-based networks primarily help with the detection of large objects \citep{Carion2020}.
From a point cloud, we predict a binary \ac{BEV} mask and a class for each detected instance. 
As shown on \autoref{fig:mask-log}, each mask represents the top-down view of an object's footprint.
These masks are complete, meaning that they capture the full footprint of the instances, as shown in \autoref{fig:mask-completion}\textcolor{red}{d}.
This effectively reformulates 3D object detection as a classification-only task, as masks are not regressed.
Moreover, it allows for object completion of arbitrary shapes.
This is due to the transformer's philosophy of reducing the number of inductive biases baked into the network \citep{Khan2022}.
Consequently, the completed shape prior is provided by the ground truth annotation in the form of masks.

\begin{figure*}[htbp]
    \vspace{7pt}
    \centering
    \includegraphics[width=\textwidth]{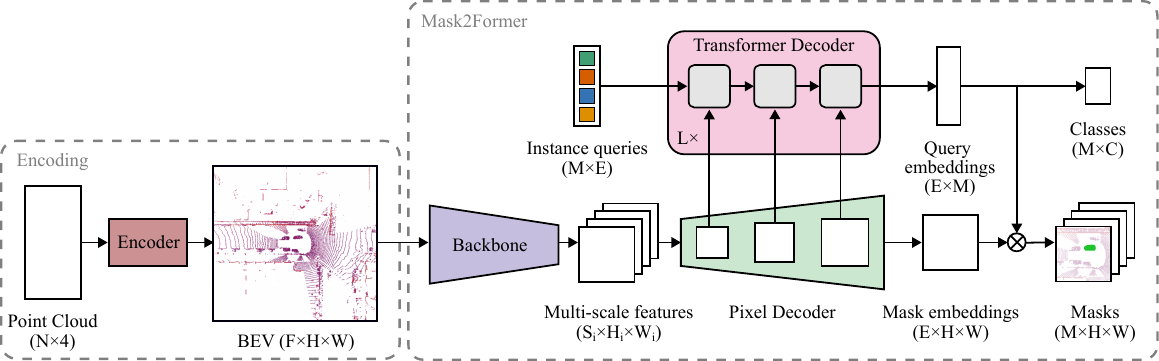}
    \caption{\textbf{MaskBEV complete architecture.}
    It has two main parts: an encoder and a mask prediction network.
    The encoder is responsible for converting a 3D point cloud into a \ac{BEV} feature map.
    Then, this feature map is fed into a Mask2Former \cite{Cheng2022} network that outputs a set of classes prediction and binary \ac{BEV} masks.
    Each class and mask pair represents a detection made by the network.
    These masks predict the footprint of each detected instance.}
    \label{fig:maskbev-archi}
    \vspace{-1.5em}
\end{figure*}

MaskBEV, as depicted in \autoref{fig:maskbev-archi}, is split into two main components: an encoder and a mask prediction module.
The encoder transforms 3D point clouds into \ac{BEV} images, allowing us to reformulate the detection task to a computer vision one.
The mask-prediction module first extracts multi-scale features and then predicts a set of up to $M$ masks, along with their class (\ie{} a binary classification of whether there is a car or not in the corresponding mask).
These two components are described in more detail below.

\subsection{Encoder}

The encoder of MaskBEV transforms point clouds into \ac{BEV} images in a similar fashion to PointPillars \citep{Lang2018}.
First, we sparsely voxelize the input 3D point cloud into a 2D grid along the ground plane in a region around the LiDAR, sampling up to 32 points per voxel.
Then, we augment each point with voxel-relative information, resulting in a ten-dimensional vector: three for the $(x, y, z)$ position, one for the distance of the voxel from the origin, three for the offset of each point from its voxel's center, and three for the offset from the arithmetic mean of all the points in the same voxel.
We also concatenate the return strength of the laser -- a value between 0 and 1 -- if available.
Finally, we generate the \ac{BEV} image by projecting all voxels using a multilayer PointNet \citep{Qi2016} followed by layer normalization.
The resulting image is of size $(F \times H \times W)$, where $F$ is the embedding dimension of voxels, $H$ and $W$ are the number of voxels along the $y$-axis and $x$-axis, respectively.

\subsection{Mask2Former}

Our mask-predicting module is based on Mask2Former~\citep{Cheng2022}.
Similarly, we extract multi-scale features from the point cloud \ac{BEV} using a Swin-T backbone \citep{Liu2021}.
We use learnable absolute positional encoding to inform the network about the voxels' position relative to the LiDAR.
The backbone produces feature maps of sizes $(S, H/2, W/2)$, $(2S, H/4, W/4)$, $(4S, H/8, W/8)$, and $(8S, H/16, W/16)$, where $S$ is the dimension of the multi-scale feature embedding.
Typically, $S$ is larger than $F$ (\ie{} the number of channels in the \ac{BEV} image) since it needs to capture the relationship between voxels.

The multi-scale feature maps produced by the backbone are upsampled using a multi-scale deformable attention transformer from \citet{Zhu2020}, ultimately producing a mask embedding of dimension $E$.
The intermediary upsampled feature maps are fed into a $L$-layers transformer decoder and queried from a set of $M$ queries tokens of dimension $E$, which in turn produce an embedding for each possible mask detection.
The query embeddings are then projected and correlated with the mask embeddings to get, respectively, a set of $M$ classes and the instance masks.
We then apply a sigmoid activation on the mask embeddings followed by a threshold of 0.5 to get the binary masks representing the footprint of detected objects.

\subsection{Mask Ground Truth Generation} \label{sec:mask-gen}

Since MaskBEV requires mask supervision to both learn detection and footprint completion, we need to transform commonly available labels into a set of binary masks.
We focus on bounding box and per-point instance labels.
For bounding boxes from KITTI, we generate \ac{BEV} mask images and draw the orthogonal projection of the bounding boxes onto the ground plane.
For semantic and instance annotations, we generate a set of instance masks from labels in SemanticKITTI by voxelizing the point cloud in a \ac{BEV} and using the voxels containing instance points to deduce the footprint of objects.
However, since only the LiDAR-facing side of objects are visible, as shown in \autoref{fig:mask-completion}\textcolor{red}{a}, this process would give incomplete \ac{BEV} masks, as depicted in \autoref{fig:mask-completion}\textcolor{red}{b}.
To remedy this problem, we build a map from sequential scans, gathering points from multiple views around static objects, as shown in \autoref{fig:mask-completion}\textcolor{red}{c}.
We then produce a \ac{BEV} mask from this map at the position of a LiDAR scan, which gives us a complete mask of objects, as illustrated in \autoref{fig:mask-completion}\textcolor{red}{d}.
We apply closing and opening morphological operations to clean the instance masks.
These operations are followed by the filtering of masks whose area is smaller than a certain threshold.
This removes instances that are never fully seen in the aggregated point cloud (\ie{} vehicles that are too far away from any single scan).
Due to this process, we limit ourselves to static vehicles in our training datasets (\ie{} parked vehicles, for masks generated from semantic labels).
This training method does not prevent MaskBEV from detecting mobile vehicles in single-scan point clouds from the test split.
When we generate a mask label for a specific LiDAR scan, we only keep masks for instances that have at least one point in the point cloud.
While the \ac{BEV} instance mask labels are generated from multiple LiDAR scans, the network only receives in input a single scan.

This generation of complete mask labels enables MaskBEV to learn how to perform \emph{object completion} in a data-driven manner.
To do so, we train MaskBEV to predict the complete footprint of vehicles, even when only parts of the objects are visible.

\section{Experiments and Results}

\subsection{Datasets}

We evaluate MaskBEV on two widely used datasets in the 3D object detection literature.
We focus our experimentation on vehicle detection because they are larger objects that could benefit from our transformer-based architecture.
We also experimentally validate the quality of the footprint completion, a key aspect of our proposed approach.

The KITTI Vision Benchmark Suite \cite{Geiger2013} is composed of a large quantity of data coming from diverse sensors installed on a standard station wagon. 
The dataset contains \num{16142} vehicles in the training set, and \num{16608} in the validation set, from all of which we generate \ac{BEV} instance masks.
For our research, we only use the data from the LiDAR (Velodyne HDL-64E) for car detection and the Inertial Navigation System (OXTS RT 3003) for localization.
KITTI contains bounding box annotations that are limited to what is visible from the image plane, meaning that vehicles behind the LiDAR are not labeled.
For this reason, the point cloud is cropped to keep the points between [0, 80] \SI{}{m} in the $x$-axis and [-40, 40] \SI{}{m} in the $y$-axis, from the sensor's reference frame.
We use the bounding box labels to generate our training \ac{BEV} instance masks.
The SemanticKITTI dataset \cite{Behley2019} provides dense point-wise annotations for the whole field-of-view of the LiDAR, for all sequences of the KITTI Vision Odometry Benchmark.
It consists of \num{19130} scans from ten sequences in the training set, totaling \num{136374} static vehicle instances.
The validation set contains \num{4071} scans all taken from a single sequence, in which there are \num{37280} instances of static vehicles.
Since we have 360$^\circ$ labels this time, the point cloud is cropped to keep the points in a \SI{40}{\m} range around the LiDAR sensor.

\subsection{Implementation Details}

Our implementation of MaskBEV, represented in \autoref{fig:maskbev-archi}, uses two PyTorch \citep{Paszke2019}
libraries, \texttt{mmdetection}\footnote{\url{https://github.com/open-mmlab/mmdetection}} and \texttt{mmdetection3d},\footnote{\url{https://github.com/open-mmlab/mmdetection3d}} respectively for their implementation of Mask2Former and PointPillars.
The points are split using voxels of size \SI{0.16}{\m}, which results in feature maps of size $H=W=500$, for both SemanticKITTI and KITTI.
The encoder is a 3-layer PointNet which outputs a \ac{BEV} representation with $F=128$ channels.
From this, a Swin-T backbone generates multi-scale features with $S=192$ channels.
The pixel decoder's output has $E=256$ channels and the transformer decoder uses $M=45$ queries.
The remaining parameters are taken from Mask2Former's default parameters~\citep{Cheng2022}.
It should be noted that we did not perform an extensive hyperparameters search.

\subsection{Training Details}

For training, we used the AdamW optimizer \citep{Loshchilov2017} with a learning rate of $1\times 10^{-4}$ and weight decay of $1\times 10^{-5}$.
We did not use a learning rate multiplier for the backbone, since we do not use pre-trained networks.
Following MaskFormer, we multiply the loss of predictions assigned to "\texttt{no\_object}" by $0.1$.
The models were trained on four Nvidia RTX A6000 GPUs, an AMD Ryzen Threadripper 3970X 32-core CPU, and 128 GB of RAM.
We train on SemanticKITTI for 21 epochs for 60 hours, and 50 epochs on KITTI in 22 hours.

For the SemanticKITTI dataset, we applied the following data augmentation: we randomly drop \SI{5}{\%} of points, apply random flipping along the $y$ axis (\ie{} a left-right swap from the vehicle's perspective), and randomly translate individual points with noise sampled from $\mathcal{N}(0, 0.2)$.
For KITTI, we needed stronger data augmentation to counter effect the lower number of vehicles per scan.
In addition to the augmentation used for SemanticKITTI, we also used the following data augmentation inspired from \citet{Lang2018}:
adding up to ten instances from other scans, making sure not to collide with existing instances;
randomly translating and rotating vehicle instances, limiting the displacement to \SI{0.25}{\m} and the rotation to 9 degrees;
randomly translating the full point cloud by a maximum of \SI{0.2}{\m}; and
randomly rotating the full point cloud by a maximum of 2.5 degrees.

\subsection{Results on SemanticKITTI}

Since most semantic and instance architectures evaluate their approach using a per-point \ac{mIoU} for SemanticKITTI, there is no direct comparison for MaskBEV with other competing approaches.
This limitation is due to our solution being the first one predicting complete \ac{BEV} masks and not a per-point prediction.
Consequently, we report usual detection metrics based on binary mask \ac{IoU}, thereby establishing a novel baseline for mask-based detection on 3D point clouds.
We evaluated MaskBEV on all vehicle instances that are visible in the point cloud no matter the amount of occlusion (\ie{} at least one point of the vehicle is present in the LiDAR scan).
In \autoref{tab:sem-kitti-results}, we present AP50, AP70, mAP and mIoU for mask predictions on SemanticKITTI's validation split.
It should be noted that the \ac{BEV} prediction task is more difficult than bounding box prediction, due to the more flexible nature of masks.

\begin{table}[htbp] %
    \vspace{7pt}
    \def\arraystretch{1.2}
    \setlength{\tabcolsep}{5.5pt}
    \caption{Results of mask detection metrics on the SemanticKITTI's validation split.}
    \centering
    \begin{tabularx}{\columnwidth}{X|c|c|c|c}
        \hline
        \textbf{Model} & \textbf{AP50} & \textbf{AP70} & \textbf{mAP} & \textbf{mIoU} \\
        \hline \hline
        MaskBEV (ours) & 77.68 & 63.11 & 48.53 & 69.93 \\
        \hline
    \end{tabularx}
    \label{tab:sem-kitti-results}
\end{table}

\subsubsection{Qualitative Results}

We show in \autoref{fig:mask-log} examples of mask predictions on a sample of point clouds from SemanticKITTI's validation split.
\autoref{fig:mask-log}\textcolor{red}{a} presents the input point cloud with the instance mask labels overlayed.
\autoref{fig:mask-log}\textcolor{red}{b} shows the network's predictions.
\autoref{fig:mask-log}\textcolor{red}{c-h} show the normalized masks before the sigmoid and the thresholding operation.
Interestingly, we can observe black outlines of cars in these raw mask predictions.
They correspond to other vehicles that the network suppressed in order to predict a single instance.
This suppression seems to indicate that MaskBEV is able to leverage the \emph{global structure} of the scene, as well as other detections, to improve performance.

\begin{figure*}[htbp] %
    \vspace{5pt}
    \centering
    \includegraphics[width=0.98\textwidth]{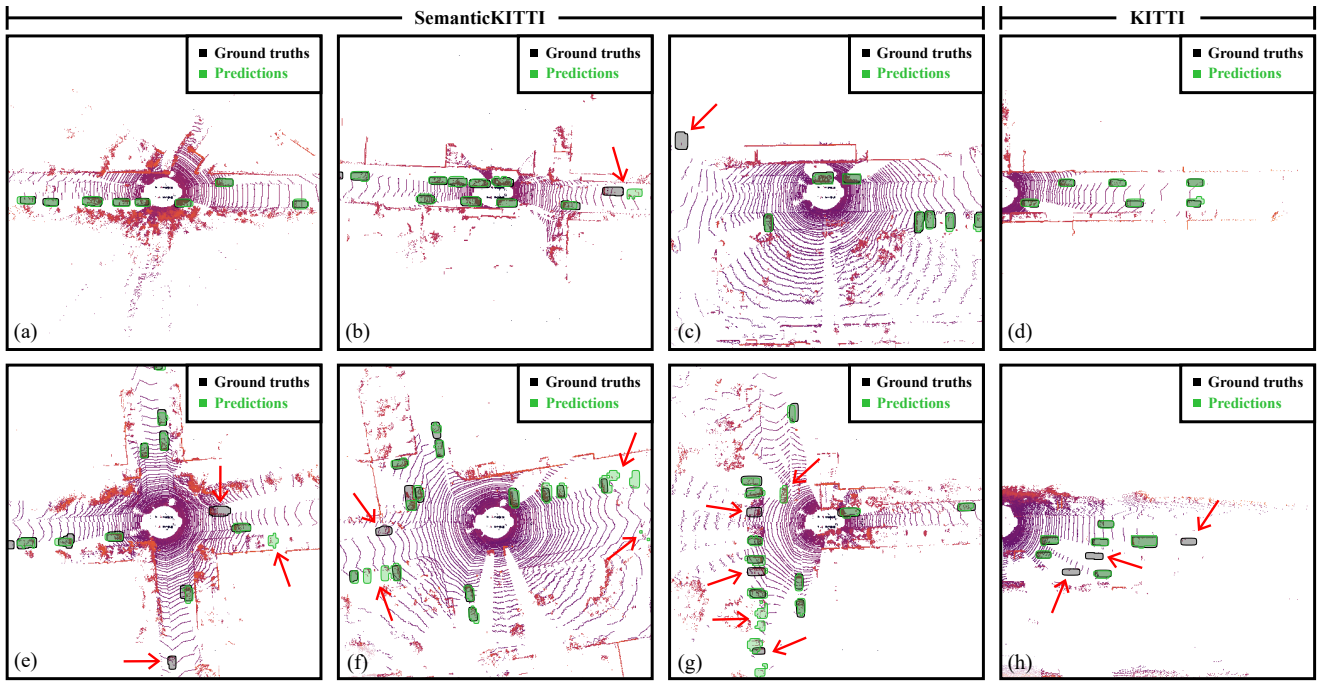}
    \caption{\textbf{MaskBEV predictions on SemanticKITTI and KITTI datasets.}
    The first three columns are samples from SemanticKITTI, and the rightmost one is from KITTI.
    The top row shows a sample of good predictions from the network.
    The bottom row displays failure cases to analyze the limitations of MaskBEV. 
    We see that the network struggles with more complex scenes such as (e), (f) and (g).
    Missed detections are emphasized by red arrows.
    Smaller ground truths (\ie{} rectangles that are too small to be vehicles), are filtered out by the process described in \autoref{sec:mask-gen}, and thus are not used for evaluation.
    }
    \label{fig:mask-quality}
    \vspace{-1.5em}
\end{figure*}

We show in \autoref{fig:mask-quality} (a-c) and (e-g) examples of predictions made by MaskBEV on SemanticKITTI. We can see that MaskBEV's predictions are more accurate on simpler point clouds, such as single roads, but struggles on more complex scene structures such as intersections.

\begin{figure}[htbp]
    \centering
    \begin{subfigure}[b]{0.5\columnwidth}
        \centering
        \includegraphics[width=\columnwidth]{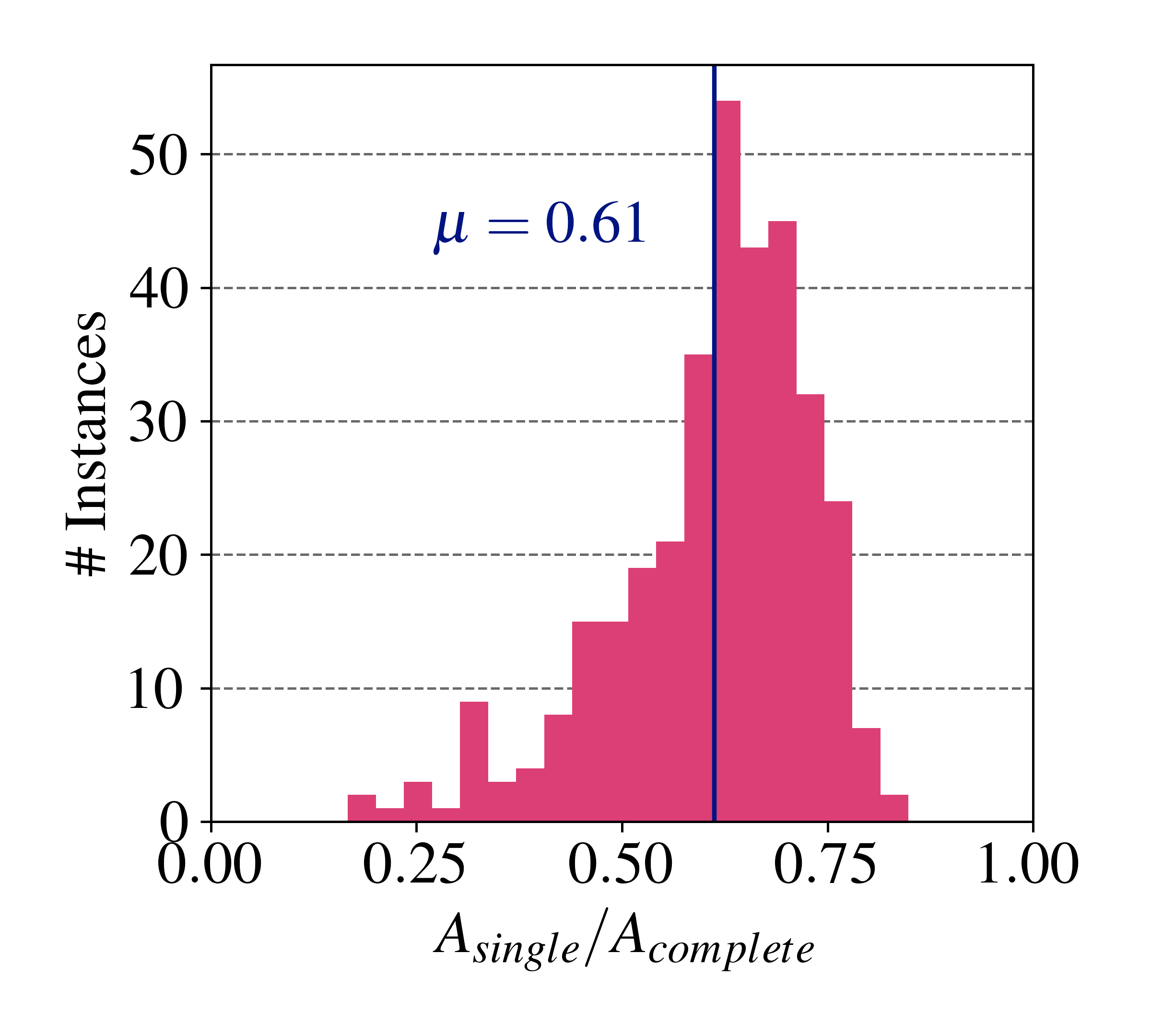}
        \vspace{-0.27in}
        \caption{}
        \label{fig:area-single}
    \end{subfigure}%
    \begin{subfigure}[b]{0.5\columnwidth}
        \centering
        \includegraphics[width=\columnwidth]{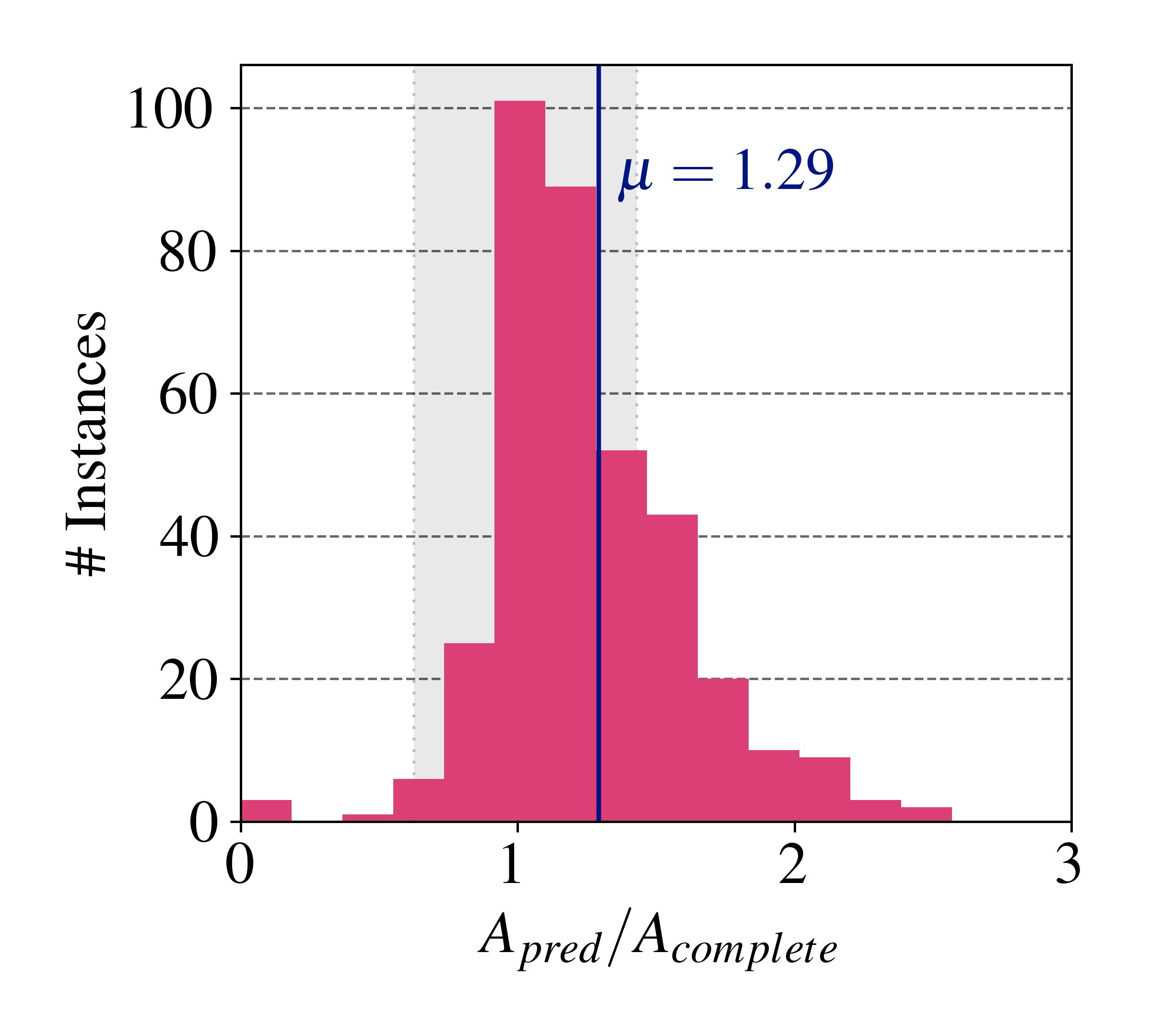}
        \vspace{-0.27in}
        \caption{}
        \label{fig:area-pred}
    \end{subfigure}
    \caption{(a) Histogram of the ratio between the area of the largest mask generated from a single scan, $A_{single}$, to the area of the complete mask of the same instance, $A_{complete}$, for SemanticKITTI's validation split.
    Lower values of ratios indicate instances that are not well captured from any single scan.
    (b) Histogram of the ratio between the area of our prediction, $A_{pred}$, to the area of the complete mask $A_{complete}$ for the same instance, for SemanticKITTI's validation split.
    Most predictions are larger than their corresponding ground truth (\ie{} ratios bigger than one), meaning that MaskBEV tends to overestimate the footprint of instances.}
    \vspace{-1.5em}
\end{figure}

\subsubsection{Mask Completion}

We provide here an analysis of the mask completion capabilities of MaskBEV.
In \autoref{fig:area-single}, we compare ground truth masks generated from a single scan and merged scans in SemanticKITTI.
To achieve this, we compute the ratio between the area of the instance masks generated from a single scan for one instance, shown in blue in \autoref{fig:mask-completion}\textcolor{red}{b}, to the area of the complete mask, outlined in gray in the same figure.
We compute this metric for all instances over all scans, and only keep the largest value per instance.
In other words, we only consider the best case, \ie{} the scan in which the vehicle is the most visible for each instance.
This ratio demonstrates how well a single scan can capture the actual footprint of objects in the ideal case.
We ignore here instances only containing a few points per scan, and thus having a low single-scan area.
The histogram in \autoref{fig:area-single} shows that most single-scan instance masks computed from the ground truth data are incomplete; they have completion ratios between 0.2 and 0.8.
Overall, the average ratio over all instances is 0.61, meaning that most single-scan masks only cover about \SI{61}{\%} of the complete footprint of the vehicles in the ideal case.
If we include the instances with a low single-scan area (\ie{} instances suffering from severe occlusion in every scan), the average ratio slightly decreases to $0.55$.
Nevertheless, our mask-generation algorithm still allows using these instances for training, since we gather points over a whole sequence and generate a complete mask.
Training on such instances could help with detecting vehicles that suffer from severe occlusion or that are far away.
If we instead consider the average ratio of all instances over all scans, and not only the ideal case, the average drops to $0.41$ with a standard deviation of $0.15$.
This ratio shows that in the average case, points on an instance only capture about \SI{40}{\%} of the complete footprint, still proving that complete masks offer a clear advantage over single-scan masks.

In \autoref{fig:area-pred}, we compare MaskBEV's predictions to the complete \ac{BEV} instance masks.
The histogram shows the ratio between the area of predictions to the area of the ground truth for each instance in SemanticKITTI's validation set.
This metric gives insight about the relative size of our prediction to the actual size of the object.
The average ratio is 1.29, meaning that MaskBEV's predictions are on average \SI{29}{\%} larger than ground truth labels.
If we compute the \ac{IoU} using only the mask areas (\ie{} assuming otherwise perfect detection) and a detection threshold of \SI{70}{\%} for the \ac{IoU}, ratios between $\frac{70}{100}=0.7$ and $\frac{100}{70}\approx 1.43$, shown as a gray zone in \autoref{fig:area-pred}, are considered as correct detections.
Thus, we can deduce that larger predictions are an important source of error for MaskBEV as the histogram is skewed towards larger values.
Larger detections are not inherently detrimental in an autonomous vehicle context, where a detected object’s enlarged footprint would result in a more conservative, safer path planning.

\subsection{Results on KITTI}

In \autoref{tab:kitti-results-70}, we evaluate MaskBEV using AP70 on the different difficulty levels defined by KITTI and \ac{mIoU} over mask predictions.
As in the case of SemanticKITTI, our evaluation metrics are computed directly on the predicted masks, making a straightforward comparison with other methods difficult.
Nevertheless, we show the performance on bounding box predictions for the state-of-the-art SE-SSD \citep{Zheng2021} approach.
These results indicate that our approach is promising, but has significant room for improvement.
Importantly, MaskBEV contains fewer inductive biases, either in its transformer backbone or the lack of rectangular anchors in the output.
Consequently, we conjecture that its performance is greatly affected by the small size of the KITTI dataset.
This phenomenon has already been seen in transformer-based neural networks in computer vision \citep{zhai2022scaling}, where it is not uncommon to pre-train these networks on billions of images.
We show qualitative examples of predictions in \autoref{fig:mask-quality} (d) and (h).

\begin{table}[htbp]
    \vspace{7pt}
    \def\arraystretch{1.2}
    \setlength{\tabcolsep}{5.5pt}
    \caption{Mask detection metrics on KITTI. 
    Results for MaskBEV are computed on mask predictions. 
    Other results are from KITTI's \ac{BEV} leaderboard for bounding boxes.}
    \centering
    \begin{tabularx}{\columnwidth}{X|ccc|c}
    \hline
    \multirow{2}{*}{\textbf{Model}} & \multicolumn{3}{c|}{\textbf{AP70 on cars}}                                        & \multirow{2}{*}{\textbf{mIoU}} \\ \cline{2-4}
                           & \multicolumn{1}{c|}{\textbf{Easy}}
                           & \multicolumn{1}{c|}{\textbf{Moderate}} 
                           & \textbf{Hard}  &  \\ 
    \hline \hline
    MaskBEV (masks, ours) & \multicolumn{1}{c|}{72.22} & \multicolumn{1}{c|}{71.02} & 54.39 & 64.79 \\ 
    \hline \hline
    SE-SSD \citep{Zheng2021} & \multicolumn{1}{c|}{95.68} & \multicolumn{1}{c|}{91.84} & 86.72 & - \\ \hline
    PV-RCNN \citep{Shi2019} & \multicolumn{1}{c|}{94.98} & \multicolumn{1}{c|}{90.65} & 86.14 & - \\ \hline
    BtcDet \citep{Xu2022} & \multicolumn{1}{c|}{92.81} & \multicolumn{1}{c|}{89.34} & 84.55 & - \\ \hline
    PointPillars \citep{Lang2018} & \multicolumn{1}{c|}{90.07} & \multicolumn{1}{c|}{86.56} & 82.81 & - \\ \hline
    \end{tabularx}
    \label{tab:kitti-results-70}
\end{table}

\section{Conclusion}

In this paper, we introduced a new 3D object detection paradigm for point clouds, based on \ac{BEV} masks instead of bounding boxes.
To do so, we developed the novel MaskBEV architecture, by combining a PointPillars encoder with Mask2Former.
MaskBEV avoids common drawbacks of anchor-based and anchor-free detectors while remaining conceptually simple.
Our method, by predicting complete \ac{BEV} masks, allows object completion while simultaneously detecting them.
This makes our approach particularly apt to deal with LiDAR scans fraught with occlusions and signal misses.
We evaluated MaskBEV on SemanticKITTI and KITTI, with competitive results.
Qualitative results indicate that MaskBEV is also able to incorporate global cues as well as the shape prior provided by the training data.
All in all, these suggest that we can be optimistic about further developments of this paradigm of mask-based object detection in point clouds.

In terms of future work, we will extend MaskBEV to other types of objects such as pedestrians, cyclists and moving vehicles.
We additionally plan on studying MaskBEV's object completion capabilities, potentially developing data augmentation techniques that take advantage of this.
We also need to further analyze how our transformer-based network can leverage both local and global scene structures.
Since our approach has less inductive biases, we can expect it to greatly benefit from larger training datasets, \eg{} KITTI-360 \citep{Liao2022}, Waymo Open Dataset \citep{Sun2020} and nuScenes \citep{Caesar2020}.
Finally, transformer-based approaches have been shown to be compatible with self-supervised learning.
As such, we plan on testing Voxel-MAE \citep{min2022voxel} with our method, in order to reduce the need for labeled data.

\printbibliography

\end{document}